\documentclass[sn-mathphys,Numbered]{sn-jnl}

\usepackage{graphicx}%
\usepackage{multirow}%
\usepackage{amsmath,amssymb,amsfonts}%
\usepackage{amsthm}%
\usepackage{mathrsfs}%
\usepackage[title]{appendix}%
\usepackage{xcolor}%
\usepackage{textcomp}%
\usepackage{manyfoot}%
\usepackage{booktabs}%
\usepackage{algorithm}%
\usepackage{algorithmicx}%
\usepackage{algpseudocode}%
\usepackage{listings}%
\usepackage[]{mdframed}
\usepackage{tcolorbox}
\newcommand{\SA}[1]{\textcolor{black}{#1}}
\newcommand{\RE}[1]{\textcolor{black}{#1}}
\usepackage{glossaries}
\usepackage{xcolor}

\theoremstyle{thmstylethree}%

\begin{document}

\title[Article Title]{{\bf MARRO: M}ulti-headed {\bf A}ttention for {\bf R}hetorical {\bf Ro}le Labeling in Legal Documents}


\author[1]{\fnm{Purbid} \sur{Bambroo}}
\author[4]{\fnm{Subinay} \sur{Adhikary}}
\author[2]{\fnm{Paheli} \sur{Bhattacharya}}
\author[3]{\fnm{Abhijnan} \sur{Chakraborty}}
\author[3]{\fnm{Saptarshi} \sur{Ghosh}}
\author[4]{\fnm{Kripabandhu} \sur{Ghosh}}

\affil[1]{\orgname{University of Utah}, \city{Salt Lake City}, \country{USA}}
\affil[2]{\orgname{Bosch Research}, \city{Bengaluru}, \country{India}}
\affil[3]{\orgname{Indian Institute of Technology Kharagpur}, \city{Kharagpur}, \country{India}}
\affil[4]{\orgname{Indian Institute of Science Education and Research Kolkata}, \city{Mohanpur}, \country{India}}


\abstract{Identification of rhetorical roles like facts, arguments, and final judgments is central to understanding a legal case document and can lend power to other downstream tasks like legal case summarization and judgment prediction. However, there are several challenges to this task. 
Legal documents are often unstructured and contain a specialized vocabulary, making it hard for conventional transformer models to understand them. Additionally, these documents run into several pages, which makes it difficult for neural models to capture the entire context at once. Lastly, there is a dearth of annotated legal documents to train deep learning models. Previous state-of-the-art approaches for this task have focused on using neural models like BiLSTM-CRF or have explored different embedding techniques to achieve decent results. While such techniques have shown that better embedding can result in improved model performance, not many models have focused on utilizing attention for learning better embeddings in sentences of a document. Additionally, it has been recently shown that advanced techniques like multi-task learning can help the models learn better representations, thereby improving performance. In this paper, we combine these two aspects by proposing a novel family of multi-task learning-based models for rhetorical role labeling, named \textsc{MARRO}, that uses transformer-inspired multi-headed attention. 
Using label shift as an auxiliary task, we show that models from the \textsc{MARRO} family achieve state-of-the-art results on two labeled datasets for rhetorical role labeling, from the Indian and UK Supreme Courts.}




\maketitle

\section{Introduction}\label{sec1}
\noindent 
Countries like India have traditionally followed a common law system, where jurisprudence is based on the doctrine of judicial precedent. In such context, any information retrieval task on the 
case documents 
such as document summarization can be complex to automate and has largely remained manual over the years.
However, as the number of cases pile up, handling legal documents manually becomes more challenging~\cite{RePEc:ags:ubzefd:18750, katju2019backlog}.
Today, legal case documents typically run into several pages, detailing all aspects of a case, where each sentence or paragraph serves a specific purpose in providing facts about the case, arguments presented by both sides, lower court rulings, etc~\cite{malik-etal-2021-ildc}. While this helps create exhaustive anecdotes for individual cases, these voluminous documents mean increased efforts for reading, parsing, and understanding them. To this end, labeling sections or sentences in a legal document can help expedite searching, summarizing, and classifying these documents. {\it Rhetorical roles} are the labels assigned to the sentences according to their semantic function in a legal case document. In a judicial setting, tasks like automated prediction~\cite{DBLP:journals/corr/abs-2201-13125, LYU2022102780}, document summarization~\cite{yousfi2010supervised, JAIN2021100388, ANAND20222141} or precedence retrieval~\cite{Thenmozhi2017ATS} can greatly benefit from {\it Rhetorical roles}, focusing only on specific sentences or paragraphs rather than having to process the entire document. \\

\noindent However, there are three major challenges in automatic rhetorical role classification:
\begin{itemize}
    
    \item Firstly, extensive domain knowledge is required to apply hand-crafted features or phrases for rhetorical role labeling. This is especially challenging since specific terms (niche terminology) differ in 
    different legal sub-domains, where even experts often disagree on the gold standard label~\cite{bhattacharya2019identification, DBLP:journals/corr/abs-2112-01836}. Moreover, different rhetorical roles can have variable lengths across documents, with often overlapping sections, making labeling of documents even more complicated. \\
    
    \item Secondly, it is hard to get labeled legal data, which is essential for creating a supervised machine-learning model for this task. \citet{bhattacharya2019identification} previously released a corpus of 50 annotated legal documents with 7 rhetorical roles, which was the largest annotated dataset for rhetorical role labeling on the Indian Supreme Court documents at the time. Recently, \citet{DBLP:journals/corr/abs-2201-13125} released a larger corpus of 354 documents from the Indian Supreme Court, annotated for the same task, but using 13 labels (12 actual rhetorical roles plus one NONE). 
    Although these datasets have contributed to benchmarking rhetorical role model performance on Indian legal documents, the overall size of the labeled corpus remains significantly limited. This underscores the need for additional annotated data to expand the existing corpus. Furthermore, the annotated labels may lack consistency when applied to legal documents from other countries. For example, in documents from the Indian Supreme Court, the ruling of the present court is mentioned at the end of the document. However, in court documents from the UK, it is seen each judge or ``lord" gives a separate judgment, instead of one consolidated judgment~\cite{article_1}.     Since documents of legal origin are so diverse based on the country of origin, there are very few labeled and good-quality legal datasets available, especially for India.\\
    
    \item \SA{Finally, legal documents are sequential, running into several pages. While models like LSTM and GRU have shown decent performance in the past~\cite{bhattacharya2019identification}, attention-based models are likely to work better in such context}. 
    \SA{Newer state-of-the-art transformer models have shown significantly better performance on some niche tasks like biomedical text mining \cite{10.1093/bioinformatics/btz682}, financial communications \cite{2020arXiv200608097Y}, and even legal document classification \cite{DBLP:journals/corr/abs-2010-02559}; however, there are limited attempts to apply them on the task of rhetorical role labeling.} This is because pretrained transformer or BERT-based models are trained on a general English vocabulary, often very different from the niche vocabulary in legal documents. Even if some models are pretrained on legal documents (LEGAL-BERT), they can not be directly applied to rhetorical role labeling. The input length for most transformer models is 512 owing to the self-attention mechanism, which scales quadratically with input length. Even newer and improved models with sliding window attention like Longformer \cite{DBLP:journals/corr/abs-2004-05150} and Lawformer \cite{XIAO202179}, a longer input of up to 4096 tokens are allowed. This is still very less in comparison to legal documents, which usually run into several pages, averaging between 400-600 sentences per document \cite{article_1}. 
\end{itemize}

\noindent Recent developments in deep learning changed the overall direction from using rule-based and hand-crafted feature-based approaches to using RNNs and 
transformers~\cite{bhattacharya2019identification}. Today, The state-of-the-art models for rhetorical labeling tasks approach it as a multi-label classification problem, where given a sequence of sentences (the entire judicial document), the model predicts the label of each sentence. Quite expectedly, and as shown in previous works \cite{bhattacharya2019identification}, the source and richness of the sentence embeddings used significantly affect the performance of the model. Consequently, much of the following works focused on using BERT or niche domain-trained LEGAL-BERT models to improve the input embeddings. While using such an approach could help improve the quality of individual sentence embeddings, legal documents are structured such that the label of each sentence (rhetorical role) greatly depends on its surrounding sentences. None of the prior works have explicitly leveraged transformer-based \cite{NIPS2017_3f5ee243} \textit{multi-headed attention} for generating contextual sentence embeddings among the sentences of a document. This approach when employed to train a model in a multitask setting helps the model learn more contextually aware embeddings, thus giving new state-of-the-art results. Furthermore, we leverage the Large Language Model (LLM) for identifying the rhetorical roles, as illustrated in Section~\ref{sub:baselines}.\\

\noindent \textbf{Research Objectives:}
\noindent In this work, we aim to showcase the effectiveness of hybrid models that combine attention mechanisms, BiLSTM-CRFs, transformers, and a multi-task learning setup to achieve state-of-the-art performance in rhetorical role labeling for legal documents. Additionally, we create and annotate a novel dataset specifically tailored for rhetorical roles in case documents from the Indian Supreme Court. Our objectives can be summarized as follows:

\begin{itemize}
\item
Contribute a new dataset for rhetorical role labeling on the Indian Supreme Court documents. 

\item 
Assess the effect of the initial embedding source (BERT, fine-tuned BERT, LEGAL-BERT, sent2vec) on the final model performance. 

\item 
Investigate whether 
adding attention for enriched context among the sentences 
can enhance model performance compared to a plain BiLSTM-CRF model. 

\item 
Explore advanced training techniques like multi-task learning to leverage the attention-enriched BiLSTM-CRF model to obtain state-of-the-art results.
\end{itemize}

\noindent \textbf{Our Contributions:} Overall, we make the following contributions through this work:

\begin{itemize}
\item 
\noindent \SA{As mentioned earlier, it is cumbersome and expensive to get labeled legal data for developing supervised models for rhetorical role labeling. We create and annotate a new dataset for rhetorical roles on case documents from the Indian Supreme Court. The dataset we introduce in this work adds 21,349 new annotated sentences over 100 new Indian court documents on top of the dataset shared by \citet{bhattacharya2019identification}, which covered 50 documents with 9,380 annotated sentences. 
The final dataset contains a total of 30,729 annotated sentences from 150 Indian court documents. 
We will make this dataset public upon acceptance of the paper. A sample of our dataset is shared for quick glance\footnote{\url{https://anonymous.4open.science/r/MARRO_Rhetorical-Role-Labeling-167D/}}. 
Furthermore, we utilize a previously released dataset of UK court documents, and benchmark our models on both datasets.\\} 


\item 
\SA{We introduce a new architecture for rhetorical role labeling. We build upon the idea that using multi-headed attention between sentences of a legal document and training in a multitask setting can give better results for rhetorical role labeling. We propose a new 
family of models named {\bf MARRO} (Multi-Headed Attention for Rhetorical Role Labeling in Legal Documents) employs transformer-inspired multi-headed attention over the embeddings learned from a sent2vec or a pretrained BERT-based model, to generate contextually rich embeddings. We train different variants of the proposed model, some using a multi-task learning setup. Using \textit{label shift} as an auxiliary task, our proposed architecture eliminates the need to explicitly fine-tune the label shift model beforehand (unlike \cite{DBLP:journals/corr/abs-2112-01836}), thus improving the overall model performance in rhetorical labeling. Extensive experiments over two separate datasets for rhetorical role labeling, over documents from the Indian and UK Supreme Courts, demonstrate improvement over the previous state-of-the-art models.}

\end{itemize}

\section{Related Work}
\noindent 
The confluence of Artificial Intelligence (AI) and the legal domain is not new. Researchers have incorporated AI and machine learning into different downstream tasks in the judiciary for decades. \citet{BENCHCAPON200397} presents a formal model for reasoning with cases that could capture and use information from previous cases. 
\citet{doi:10.1080/08839510490279861} showed causal inferencing (whether the evidence in the case was caused by the accused's criminal activity or by some other cause) can be modeled using explanatory coherence and Bayesian networks. As the focus on legal AI increased, more research started on tasks like legal document summarization and automated prediction (e.g., of case judgements). One of the earliest works on summarizing aided by rhetorical roles was presented in \cite{teufel-moens-2002-articles} where the authors restored the discourse context of the document by adding the rhetorical status to each sentence in scientific texts. This paved the way for the following works  \cite{grover2003summarising} where automatic linguistic annotation was performed using XML-based tools, for a small sample set, to explore correlations between linguistic features and argumentative roles. AI found its application in other downstream tasks such as judgment prediction \cite{SHAIKH20202393, XIAO202179, LYU2022102780}, legal document similarity computation~\cite{inproceedings, BHATTACHARYA2022103069, BI2022109046}, catchphrase identification~\cite{galgani2012towards, 10.1145/3132847.3133102}, legal query understanding~\cite{shankar2018deep, 10.1007/978-3-030-23281-8_4}, legal attribute extraction~\cite{adhikary2024case,adhikary2023towards}, legal charge prediction~\cite{10.1007/978-3-030-30490-4_20, 10.1145/3511808.3557379}, legal document classification~\cite{DEARRIBAPEREZ202210180, SONG2022101718}, legal coreference resolution~\cite{JI2020102365}, and legal case retrieval~\cite{LIU2022103051}. 

Although most of the recent works in legal NLP have been based on deep learning, reinforcement learning, and non-deep learning-based approaches have also seen a fair share of success. ~\citet{CHEN2022102798} showed using domain concepts as features, and a random forest classifier significantly outperforms deep learning systems (built using pretrained word embeddings) on a text classification task. The said experiments were performed on a set of 30,000 full case documents in 50 different categories. ~\citet{CHEN2022102798} presented a novel reinforcement learning-based approach for Legal Judgment Prediction (LJP). This work suggests current approaches for LJP suffer from ambiguous fact descriptors and misleading articles with highly similar TF-IDF representations. The work proposes the use of four types of discriminative criminal elements while using a reinforcement learning-based extractor to accurately locate elements for different cases. Rhetorical role classification has been shown to provide power to the above-mentioned downstream tasks, hence a superior model for the prediction of rhetorical role can be of significant help to advocates, legal practitioners, and law schools alike. Most of the models previously proposed for rhetorical role classification can be divided into either (a) Non-deep learning approaches (Handcrafted feature models), or (b) Deep learning approaches (BiLSTM or Transformer based). We now take a look at both these categories below.

\vspace{2mm}
\noindent \textbf{Non-deep learning approaches}: ~\cite{Moens1999AbstractingOL} proposed clustering for abstracting court decisions for criminal case documents. Most of the initial approaches for semantically segmenting sections of legal documents were based around helping the summarization task \cite{grover2003summarising}. 
Using judgments from the House of Lords (UK), the authors divided the texts into 3 basic rhetorical schemes: Background, Case, and Own. These schemes then contained the actual rhetorical roles or sentence labels for the text. This would then be used to create a correlation between sentence type (based on labels) and verb group properties. Subsequent works in \cite{articleLETSUM} focused on identifying thematic structures in a legal text, to find the argumentative themes of the judgment. By categorizing the sentences into 4 themes: namely Introduction, Context, Juridical Analysis, and Conclusion, the authors then used sentences from each theme to contribute to the final summary (based on relevance). The use of handcrafted features was shown to be effective by \cite{articleCRF}, where the authors used conditional random field (CRF) models to semantically segment sentences into one of seven rhetorical roles. \cite{DBLP:journals/corr/abs-1710-09306} employed mean probability ensemble systems to predict the case ruling.  

\noindent \textbf{Deep learning approaches}: Over time with the advent of supervised learning models and hardware resources to aid such models, machine learning began taking over the legal AI space, with most state-of-the-art models using deep learning models for rhetorical role prediction. Recent developments in deep learning changed the overall direction from using rule-based and hand-crafted feature-based approaches to using RNNs and transformers \cite{Chalkidis2017ADL, bhattacharya2019identification, 10.1007/978-3-030-23281-8_4}. \cite{bhattacharya2019identification} released a model based on bidirectional LSTM followed by a Conditional Random Field (BiLSTM-CRF) that achieved state-of-the-art performance on multiple datasets, where the CRF learns the transition scores on probabilities generated by the LSTM model. Along with the model, the authors also released an annotated dataset of 50 documents from the Indian Supreme Court, with each sentence tagged with one of seven rhetorical roles. This was the largest dataset for rhetorical role classification on Indian Supreme Court documents at the time. \cite{MUMCUOGLU2021102684} exhibited the efficacy of NLP techniques on Turkish court documents using models like Decision Trees, Random Forests, LSTMs, and attention for judgment prediction. A fine-grain annotation approach for Named Entity Recognition (NER) was explored by \cite{CORREIA2022102794}, where 76 law students annotated a corpus of 594 decisions from the Brazilian Supreme Court (STF). They used four coarser labels (named entities), followed by twenty-four fine-grained labels. Given the limited amount of annotated data present in the legal domain for machine learning, the application of multi-task models has also been on the rise in the last few years. \cite{inproceedings} particularly showed how multi-task learning models can aid better deep learning for translation, summarization, and multi-label classification, for a German dataset. \cite{DBLP:journals/corr/abs-2112-01836} proposed a multi-task learning approach for rhetorical role labeling, where prediction of label shift in legal documents is chosen as an auxiliary task to supplement the main task. \cite{YANG2022107960} proposed a multi-task learning-based legal judgment prediction framework that jointly models several subtasks. This framework, called ``MVE-FLK" uses a multi-view encoder to fuse legal keywords. Other more recent works like \cite{10.1145/3511808.3557273} have formalized civil case judgment prediction (CCJP) as a multi-task learning problem by constructing three tasks (predictions on civil causes, law articles, and the final judgment on each plea), using the results of each subtask to make the final judgment prediction. \cite{belfathi2023harnessing} leveraged the Large Language Model (e.g., GPT-3.5) to predict rhetorical roles. In addition, LLMs are utilized in different legal tasks~\cite{DBLP:journals/corr/abs-2402-18502, DBLP:conf/emnlp/VatsZDSBNGRG23} as well.

\noindent Our work comes close to and builds upon previous works by \cite{bhattacharya2019identification} and \cite{DBLP:journals/corr/abs-2112-01836}. Like \cite{bhattacharya2019identification}, we employ the use of BiLSTM-CRF architecture in our model (along with other transformer embeddings and multi-headed attention). Our work also shows the best performance when the models are trained in a multi-task learning setup, much like \cite{DBLP:journals/corr/abs-2112-01836}, taking label shift prediction as an auxiliary task. \SA{\cite{belfathi2023enhancing,de2023irit_iris_a} leveraged BERT-based model for identifying legal rhetorical roles from a document.}
In contrast, this paper proposes a novel approach of using transformer models and attention in conjunction, trained in a multi-task architecture to generate contextually aware embeddings for the rhetorical role classification task. In contrast to~\cite{DBLP:journals/corr/abs-2112-01836}, we propose an improved architecture that eliminates the need to explicitly fine-tune the label shift model beforehand, thus saving time and compute resources, while improving the model performance. 

\section{Dataset}
\label{sec:dataset}

\noindent In this work, we consider the following set of {\it seven} rhetorical roles introduced  in~\cite{bhattacharya2019identification}:
\begin{enumerate}
    \item \textbf{Facts (FAC):} This label refers to the facts pertinent to the case. 
     \item \textbf{Ruling by Lower Court (RLC):} 
    This label refers to the verdicts of lower courts (Trial Courts, High Courts, and Tribunals) and the ratio behind these judgments.
    \item \textbf{Argument (ARG):} This label refers to the arguments of the contending parties.
    \item \textbf{Ratio of the decision (RATIO):} This label refers to the application of the law along with the rationale on the points argued in the case.
    
    \item \textbf{Statute (STA):} Established laws the court refers to, usually coming from Acts, Articles, Rules, Orders, Quotations directly from the bare act, Notices, etc.
    \item \textbf{Precedent (PRE):} This label refers to the prior cases cited as a justification or analogy in the context of the current case.
    \item \textbf{Ruling by the Present Court (RPC):} This label refers to the final judgment/ decision of the court for the case.
   
\end{enumerate}

\vspace{1mm}
\noindent 
\textbf{Selection of documents:} 
In this work, we consider legal case documents from the Indian Supreme Court (IN) and the UK Supreme Court (UK).

The dataset of Indian case documents (which we call \textbf{$D_{IN}$}) consists of 150 judgments from the Supreme Court of India. 
This dataset contains a set of newly annotated 100 documents that we introduce in this work, along with 50 annotated documents previously released by~\cite{bhattacharya2019identification}.  The process of annotation, labels used, and the source of documents are identical for both these works, hence we present our results on the larger set of 150 documents, instead of just the newly annotated 100 documents. 
The $D_{IN}$ dataset contains 
$30,729$ sentences in total, each annotated with one of the rhetorical labels stated above. 

For the $D_{IN}$ dataset, the final 150 documents are sampled from the following five categories of court documents (30 documents from each category):
(1)~Criminal, (2)~Land \& Property, (3)~Constitutional, (4)~Labour \& Industrial, and (5)~Intellectual Property Rights.

The second dataset we use in this work comes from the UK Supreme Court case documents (which we call \textbf{$D_{UK}$}) previously released by~\cite{article_1}, containing $50$ documents collected from the official UK Supreme Court website\footnote{\url{ https://www.supremecourt.uk}}. 
There are a total of $18,155$ sentences in the $D_{UK}$ dataset.
Unlike the $D_{IN}$ dataset, the documents in $D_{UK}$ were sampled directly from the official website of the UK Supreme Court, where no subcategories are given (as stated in the work~\cite{article_1}).

The datasets $D_{IN}$ and $D_{UK}$ have no overlap in them hence results are reported separately for both. We found that the UK Supreme Court documents are 56\% lengthier than the documents of the Indian Supreme Court, with the average Indian Supreme Court document in our dataset being 205 sentences. Whereas, the UK Supreme Court documents had an average sentence length of 363 sentences. \RE{Furthermore, the sentences can be any length, as this is an intentional design choice. Often, the last few sentences are where the judgment is pronounced and it can be the case the sentence is only a few tokens long with words like ``guilty'', ``not guilty'', ``dismissed'', etc. We include sentences of all lengths in our analysis to ensure no critical information is lost.}

\vspace{2mm}
\noindent \textbf{Annotating the documents:} The annotation is carried out by three recent LLB graduates from the Rajiv Gandhi School of Intellectual Property Law, India, one of the most reputed Law schools in India. The experts were asked to label each sentence into one of the seven rhetorical roles stated above. We followed a systematic way of annotation similar to~\cite{DBLP:journals/ail/ShulayevaSW17} and~\cite{Wyner2013ACS}.  Each sentence was labeled with one rhetorical role and no sentence was not assigned any label. Since rhetorical role labels are often subjective, we take a look at inter-annotator agreement scores as a measure of agreement among the annotators. \\

\noindent \textbf{Computing inter-annotator agreement}: As noted in~\cite{Wyner2013ACS}, aggregated pairwise Precision, Recall, and F1 Score are more suitable measures for inter-annotator agreement than measures like Kappa score \cite{artstein-poesio-2008-survey}. Following the same line, we compute these pairwise inter-annotator agreement measures. Since we have three annotators ($A_1$, $A_2$, and $A_3$), we compute three sets of pairwise inter-annotator agreements ($A_1$, $A_2$), ($A_2$, $A_3$), ($A_1$, $A_3$), and then take the average of the three sets. It is to be noted that the IAA scores are document-wise agreement scores. For each annotator pair ($A_i$, $A_j$), the labels annotated by $A_i$ are considered {\it key annotations} while the labels annotated by $A_j$ are called {\it response annotations}. The Precision, Recall, and F-score computations are done as follows:\\
\begin{displaymath}
Precision = \frac{|\left \{  \text{labels given by Ai }\right \}\bigcap \left \{ \text{labels given by Aj } \right \}|}{| \{ \text{labels given by Aj}\}|}
\end{displaymath}

\begin{displaymath}
Recall = \frac{|\left \{  \text{labels given by Ai }\right \}\bigcap \left \{ \text{labels given by Aj } \right \}|}{| \{ \text{labels given by Ai}\}|}
\end{displaymath}

\begin{displaymath}
F-Score= \frac{\text{2 * Precision * Recall}}{\text{Precision + Recall}}
\end{displaymath}

\noindent These document-wise IAA scores are averaged over the entire dataset (150 documents and 50 documents for India and UK court documents respectively) and for each annotator pair to obtain the final F scores for the entire dataset.The inter-annotator agreement was $0.921$ for $D_{IN}$ and $0.915$ for $D_{UK}$. This indicates high overall agreement in general.
Labels PRE and Ratio exhibited significant disagreements ($>= 0.3\%$) in inter-annotation computations in both datasets, showing even experts disagreed on the gold standard labels for these rhetorical roles.\\

\noindent \textbf{Curation of gold standard labels}: We consider the {\it majority opinion} of the annotators for a particular sentence as the gold standard label of that sentence. Table~\ref{tab:labels_count} gives the statistics of the dataset. As can be seen, the majority of sentences in the documents are ratio or facts, while argument, statute, RLC, and RLC together make up less than 15\% of the data. 

\begin{table}
\centering
\resizebox{0.99\textwidth}{!}{
\begin{tabular}{||p{3cm}||p{3cm}|p{3cm}|| }
\hline

Rhetorical Role & $D_{IN}$ & $D_{UK}$\\ \hline
\hline
FAC &  6,342 (20.63\%)  & 2,610 (14.38\%) \\
\hline
ARG &  1,638 (5.33\%)  &  691  (3.81\%) \\
\hline
RATIO & 17,322 (56.37\%)  &  11,231  (61.86\%) \\
\hline
STA & 1,319 (4.29\%) &  1,264  (6.96\%) \\
\hline
PRE &  3,047 (9.91\%)  & 1,526  (8.41\%) \\
\hline
RPC & 546 (1.77\%) &  297 (1.64\%) \\
\hline
RLC & 515 (1.67\%)  &  536 (2.95\%) \\
\hline
\end{tabular}}
\caption{Table showing the number of sentences having the corresponding rhetorical role labels in both datasets.}
\label{tab:labels_count}
\end{table}

\section{Methods for Rhetorical Role Labeling}
\noindent
In this section, we discuss the previous implementations of the rhetorical roles classification problem in greater detail. While most of previous works have relied heavily on using a BiLSTM-CRF architecture, of late there has been a surge in the incorporation of transformers and techniques like multi-task learning. Upon discussion of existing baseline models, we delineate our proposed family of models \textbf{MARRO, a set of four models} that employ BiLSTM-CRF with attention. All members/variants of the MARRO family mentioned below, differ in training style, and the embeddings used, but follow the same basic architecture of a BiLSTM-CRF with attention layers.  Lastly, with the help of a detailed architecture, we discuss how we employ multi-task learning to train a more efficient and faster-trained MARRO model.

\subsection{Baseline Models} \label{sub:baselines}
\noindent Existing works involving neural models for rhetorical role classification 
have mostly used Bidirectional LSTM with Conditional Random Field (BiLSTM-CRF) architecture with some modification. Such models have shown improvement over older models that employed handcrafted features and CRF. Such comparisons of handcrafted models and new age neural models have been made at length in previous works \cite{article_1}. Hence, we keep handcrafted feature models out of our discussion and focus on newer neural models for baseline comparison. Of late, the models that have shown the best performance on rhetorical role labeling tasks employ some specialized embeddings (transformer-based or otherwise) along with the conventional BiLSTM-CRF. This has been complemented by techniques like multi-task learning where an auxiliary task is defined to help the model generalize better on the main task (rhetorical role labeling). We compare our model's performance against previous works in rhetorical role labeling, specifically the following four baseline models:\\

\noindent
\textbf{1) Hierarchical BiLSTM-CRF \cite{bhattacharya2019identification}:} ~\cite{bhattacharya2019identification} suggested use of pretrained sent2vec embeddings with a Hierarchical BiLSTM-CRF model. While BiLSTM-CRF models had been previously used for multi-label classification tasks, the authors showed using pretrained embeddings from a sent2vec model (specifically trained on legal texts from the Indian Supreme Court), produced state-of-the-art results, outperforming previous baseline handcrafted features' models. They additionally perform a comparison between the said model when pretrained sent2vec embeddings are used versus when input embeddings are randomly initialized. The authors also released a dataset for rhetorical role labeling on 50 documents from the Indian Supreme Court. We introduce an additional 100 new documents to this dataset, creating a total of 150 documents that we use as a dataset in our work.\\
\begin{figure}[t]
\begin{tcolorbox}
\footnotesize
Given a segment of a legal case document as enclosed within angular brackets, enlist the likely label that apply for this segment.\\

Following are the labels along with their descriptions.\\
1. ``\textcolor{blue}{Facts (FAC)}'': ``\textcolor{blue}{This label refers to the facts pertinent to the case.}'' \\  
2. ``\textcolor{blue}{Argument (ARG)}'': ``\textcolor{blue}{ This label refers to the arguments of the contending parties.}''  \\
... \\

Instruction: Learn from the examples provided. Avoid generating fabricated or invalid label.\\

Input segment: $\langle$[INPUT]$\rangle$
\end{tcolorbox}
\caption{Descriptions of labels are substituted in the corresponding slots of the template shown in Figure~\ref{tab:prompt_template}. 
}
\label{fig:zero-shot}
\end{figure}

\noindent
\textbf{2) LEGAL-BERT-BiLSTM-CRF \cite{article_1}:} Building on their previous work, \cite{article_1} released a novel LEGAL-BERT model for rhetorical role classification. This was developed to generate better embeddings for case documents which were then fed into a BiLSTM-CRF, which showed considerable performance improvement. In this work, the authors also release a new dataset of 50 documents from the UK Supreme Court, carrying out the experiment on this dataset and the previously released dataset of 50 documents from the Indian Supreme Court. In their work, the authors suggest two state-of-the-art models Hier-BiLSTM-CRF and TF-BiLSTM-CRF for the Indian and UK court documents respectively. Experiments performed using different models like BERT and LEGAL-BERT for generating the input embeddings suggested using LEGAL-BERT generated embeddings gave the best results on the UK dataset. However, for the Indian court case documents, the sent2vec initialized embeddings still performed better than transformer-based embeddings.  \\

\noindent
\textbf{3) sciBERT-HSLN \cite{DBLP:journals/corr/abs-2201-13125}:} Kalamkar et al.~\cite{DBLP:journals/corr/abs-2201-13125} used the sciBERT-HSLN model for the task, that aims at using BERT \cite{devlin2018bert} embeddings with a BiLSTM-attention-pooling-CRF architecture. \\

\noindent
\textbf{4) MTL (BERT-SC) \cite{DBLP:journals/corr/abs-2112-01836}:} Malik et al.~\cite{DBLP:journals/corr/abs-2112-01836} suggested using an auxiliary task like label shift prediction can help a model like BiLSTM-CRF learn better embeddings. Using a multitask setting to train the said model, they can achieve state-of-the-art results on their dataset, while showing the effectiveness of the said model on downstream tasks like judgment prediction. For the MTL model, they used label shift as an auxiliary task. It is to be noted, that both the auxiliary task model and rhetorical role prediction model use the same BiLSTM-CRF architecture in conjunction, but with different input embeddings. The authors create a set of 150 documents (different from our work) annotated for rhetorical role classification. Unlike~\cite{bhattacharya2019identification} however, they use 13 different rhetorical roles instead of 7. \\

\begin{figure}[t]
\begin{tcolorbox}
\footnotesize
Given a segment of a legal case document as enclosed within angular brackets, enlist the likely label that apply to this segment.\\

Following are the labels along with their descriptions.\\
1. ``[LABEL]'': ``[Description]'' \\  
2. ``[LABEL]'': ``[Description]''  \\
... \\

Following is a list of example text for each label:\\
1. The text ``[SPAN]'' is of type ``[LABEL]'' \\   
2. The text ``[SPAN]'' is of type ``[LABEL]'' \\   
...\\

Instruction: Learn from the examples provided. Avoid generating fabricated or invalid label.\\

Input segment: $\langle$[INPUT]$\rangle$
\end{tcolorbox}
\caption{The structure of the prompt used in our ICL experiments. [X] represents a variable that is to be substituted for its value. Figure~\ref{fig:prompt_values} shows a concrete instance of this template with substituted values for the context variables.
}
\label{tab:prompt_template}
\end{figure}

\begin{figure}[t]
\begin{tcolorbox}
\footnotesize
Given a segment of a legal case document as enclosed within angular brackets, enlist the likely label that apply for this segment.\\

Following are the labels along with their descriptions.\\
1. ``\textcolor{blue}{Facts (FAC)}'': ``\textcolor{blue}{This label refers to the facts pertinent to the case.}'' \\  
2. ``\textcolor{blue}{Argument (ARG)}'': ``\textcolor{blue}{ This label refers to the arguments of the contending parties.}''  \\
... \\

Following is a list of example text for each label:\\

The text ``\textcolor{blue}{ The appellant Kashmira Singh has been convicted of the murder of one Ramesh, a small boy aged five, and has been sentenced to death.}'' is of type ``\textcolor{blue}{Facts (FAC)}''.\\
2. The text ``\textcolor{blue}{ The points against the appellant are (1)that he had a motive and that he said he would be revenged, (2) that he was ...}'' is of type ``\textcolor{blue}{Argument (ARG)}''.\\
...\\

Instruction: Learn from the examples provided. Avoid generating fabricated or invalid label.\\

Input segment: $\langle$[INPUT]$\rangle$
\end{tcolorbox}
\caption{Values of tags, descriptions and annotated spans substituted in the corresponding slots of the template shown in Figure~\ref{tab:prompt_template}. 
}
\label{fig:prompt_values}
\end{figure}

\noindent
\textbf{5)LLM in zero-shot mode:} Apart from training a deep learning model from the scratch or finetuning  a pre-trainined model, In-context learning (ICL) is a new paradigm using the language model (LM) to perform many NLP tasks \cite{brown2020language, zhao2021calibrate}, including entity recognition~\cite{ma2016label}, relation extraction~\cite{levy2017zero}, document extraction~\cite{pushp2017train} without parameters updates. In ICL, zero-shot learning is an approach where LM can directly output the prediction of a given test input without any training. In this approach, we provide only the label description in the prompt (as shown in Figure~\ref{fig:zero-shot}), and based on that information, LM predicts the label of the corresponding Input. We employ the Gemini-pro\footnote{\url{https://deepmind.google/technologies/gemini/}} model in this work,  since this model has been utilized in legal NLP tasks~\cite{nigam2024legal,DBLP:journals/corr/abs-2402-18502}.


\noindent
\textbf{6)LLM in few-shot mode:} To elevate the performance of ICL, few examples are provided to the LM, namely called \textit{few-shot learning} \cite{brown2020language,zhang2022active}. We provide an example for each label, and those examples are selected randomly from the training set. Exemplars are formed as sentences and labels, as shown in Figure~\ref{fig:prompt_values}. Eventually, Gemini-pro model predicts the label for the corresponding label, and the results are shown in Table~\ref{table:results_all_models}.

\subsection{Our Proposed Method} 
\noindent We model the task of rhetorical role prediction as a multi-class classification problem. \RE{Previous works \cite{article_1} have shown input embeddings from the LEGAL-BERT model work better with the $D_{UK}$ dataset and for the $D_{IN}$ dataset, sent2vec embeddings work better, as LEGAL-BERT training data has better overlap with $D_{UK}$ training set in terms of legal terminology.} Additionally, \cite{DBLP:journals/corr/abs-2112-01836} showed the effectiveness of the multi-task learning model for the rhetorical role classification tasks. Following these results, we propose two models from the MARRO family for the main task. MARRO \textsubscript{base} (uses sent2vec embeddings from a model trained on 53,000 Indian supreme court documents, provided publicly by~\cite{bhattacharya2019identification}), and TF-MARRO (uses LEGAL-BERT-SMALL \cite{DBLP:journals/corr/abs-2010-02559} embeddings). 
We further propose training these models in a multitask setting,  using an auxiliary task, to give up the MTL-based members of the MARRO family, namely MTL-MARRO (training MARRO \textsubscript{base} in an MTL setting) and MTL-TF-MARRO (training TF-MARRO in an MTL setting). A comparison of our proposed models and baseline models is given in Table~\ref{table:models_desc}.

\begin{figure}[t]
    \centering
    \includegraphics[width=0.5\textwidth]{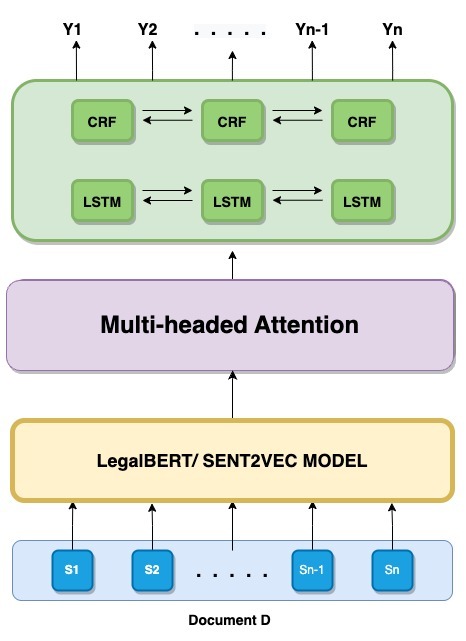}
\label{fig:Base Models}
\caption{The Model Architecture for MARRO\textsubscript{base} and TF-MARRO}
\end{figure}
\begin{figure}[t]
    \centering
    \includegraphics[width=1\textwidth]{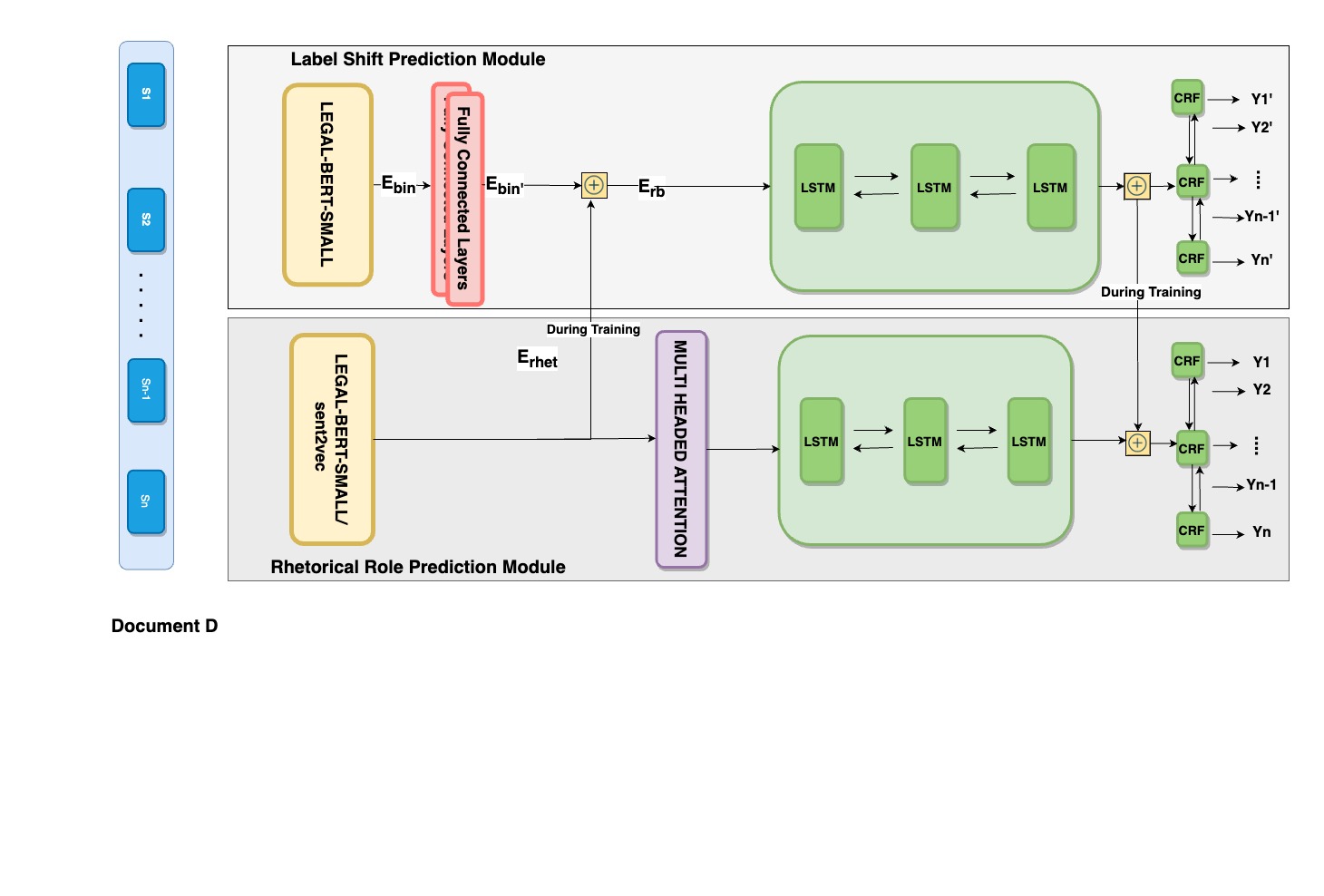}
    
\vspace{-20mm} 

\label{fig:The Model Architecture}
\caption{The Model Architecture for MTL-based MARRO models (MTL-MARRO and MTL-TF-MARRO). The embeddings are shared between the label shift prediction module and the rhetorical role prediction module during the training. While inferencing or prediction, the rhetorical role prediction module works alone.}
\end{figure}

\vspace{2mm}
\noindent \textbf{1. MARRO\textsubscript{base}}: As a basic model, we take inspiration from previously proposed BiLSTM-CRF, but this time adding transformer style multi-headed attention. Unless specified otherwise, anytime MARRO is used to refer to a single model rather than a family of models, we are referring to MARRO\textsubscript{base}.

The core idea of the attention mechanism is to focus on the relevant parts of input sentences (surrounding sentences) for each input (sentences that make up a document). The most commonly known form of attention was introduced by \cite{article} and found applications in Neural Machine Translation. Building on this concept, self-attention was introduced where each hidden state attends to the previous hidden state of the RNN. Finally, the multi-headed self-attention, popularised in language modeling using transformer \cite{NIPS2017_3f5ee243} used multiple ``heads" for self-attention calculation, where each head would focus on different aspects of language.

The application of multi-headed attention is significant in a setting like ours, where the label of each sentence depends on the label of sentences around it. However, the length of each document makes it difficult for BiLSTM-CRF to perform well or for the entire document to be fed into a transformer-based model at once. Using pretrained sentence embeddings from a sent2vec model, we can generate judiciary-pertinent embeddings for sentences in a document. The sent2vec embedding models are trained separately for both $D_{UK}$ and $D_{IN}$ datasets. For the $D_{IN}$ dataset, the model is trained on 53,000 documents from the Indian Supreme Court documents, and for the $D_{UK}$ dataset, the model is trained on 700 documents from the UK Supreme Court. For both these models, this training set is separate from the set of documents being used for rhetorical role classification. Now using multi-headed attention over these embeddings before feeding them into the BiLSTM makes sure each sentence's representation gets enriched by the surrounding context. These contextually aware embeddings are then input into the BiLSTM-CRF model like the previous works.

\vspace{2mm} \noindent 
\textbf{2. TF-MARRO}: This model works similarly to MARRO\textsubscript{base}, except the initial embeddings come from a transformer-based model and not a pretrained sent2vec model. Previous works have used input embeddings from transformer-based models, be it BERT or LEGAL-BERT. \SA{Each sentence in a document is fed into a BERT or LEGAL-BERT model one by one to get the last hidden layer representation from the model.} This embedding represents the entire input sentence. It is to be noted that while adding a transformer model, initial input embeddings can be better for certain datasets. The embedding model is fine-tuned/trained during the TF-MARRO training (\textcolor{black}{only the last two model layers are kept trainable}), leading to a significantly higher training time, as compared to MARRO\textsubscript{base}. Figure 1 shows the architecture for MARRO\textsubscript{base} and TF-MARRO.

\vspace{2mm} \noindent
\textbf{3. MTL-based MARRO models}: Legal texts follow a semantic structure, with the label shift in consecutive sentences being rare. \cite{DBLP:journals/corr/abs-2112-01836} note in 88\% of cases, the label of a sentence is the same as the following sentence's label. This means using label-shift prediction could help as an auxiliary task to learn better and more robust embeddings for the rhetorical role task. \SA{Given two consecutive sentences S\textsubscript{i} \& S\textsubscript{i+1} with rhetorical roles Y\textsubscript{i} and Y\textsubscript{i+1} in document $D \in D_{IN}$ or $D_{UK}$, a label shift is said to have occurred at position i if Y\textsubscript{i}!=Y\textsubscript{i+1}.} Using an auxiliary task of label shift prediction, we can further contextualize the BiLSTM embeddings for the main. This improved BiLSTM-CRF model with attention-based embeddings is trained. 
The entire architecture is shown in Figure 2.

We train two variants using a multi-task learning setup, one using pretrained sent2vec embeddings (building on MARRO\textsubscript{base}) and another using LEGAL-BERT embeddings (building on TF-MARRO). For both these models, multi-task learning is used with a label-shift prediction as an auxiliary task. The difference lies in the source of embeddings used. The following paragraphs explain in detail the two modules that make up the multi-task learning architecture (auxiliary task module and main task model):\\

\vspace{2mm} \noindent
\textbf{3a. Label Shift Prediction Module:} \noindent As opposed to previous works that used static embeddings from a fine-tuned model, we hypothesize that input embeddings for the auxiliary task get better as the training progresses. We hence use an off-the-shelf BERT-based model (e.g., LEGAL-BERT-SMALL \cite{DBLP:journals/corr/abs-2010-02559}) for the auxiliary task model, which gives the initial input embedding for the sentence $S_{i}$. 
\SA{This input embedding ($S_{i}^{'}$) is trained using two fully connected linear layers and then combined with input from the main task (rhetorical role prediction) through concatenation of embeddings obtained from the label shift submodule and the semantic segmentation/rhetorical role labeling submodule.}
Hence the initial input to the label shift prediction model can be defined as:
\[E\textsubscript{bin} = S\textsubscript{i}' + S_{i+1}{'}\]

Once this embedding is obtained from the BERT-based model, it is passed through two fully connected layers, and then output is concatenated with the initial embeddings from the rhetorical role prediction module as:
\[E\textsubscript{rb} = E\textsubscript{bin'} + E\textsubscript{rhet} \] 
\noindent This embedding E\textsubscript{rb} is then passed through a BiLSTM-CRF, to get the binary classification label. It should be noted the performance improvement (as compared to previous label-shift prediction models) comes from using a BERT model that is progressively fine-tuned and adding fully connected layers in the middle. \\

\vspace{2mm} \noindent
\textbf{3b. Rhetorical Role Prediction Module} \noindent For the main task we use a BiLSTM-CRF with multi-headed attention for multi-class classification. This module uses the exact same architecture as the MARRO\textsubscript{base} or TF-MARRO, depending on the source for the initial embeddings. \SA{The outputs from the BiLSTM of the label shift prediction module and the BiLSTM of the rhetorical role prediction module are concatenated. This output is then passed to the CRF, which gives us the output Y\textsubscript{i} for the rhetorical role prediction.} \\

\textbf{MTL-MARRO}: This variant of MARRO\textsubscript{base} uses pretrained sent2vec embeddings and the model is trained in a multi-task learning setup, as explained above.\\ 

\begin{table}[t]
\resizebox{1\textwidth}{!}{
\begin{tabular}{||p{5cm}|p{10.3cm}|| }
\hline
\textbf{Model}  & \textbf{Description}  \\ \hline
\hline
\begin{tabular}{l}Hierarchical BiLSTM-CRF~\cite{bhattacharya2019identification}\end{tabular} 
& Uses pretrained sent2vec embeddings with a BiLSTM-CRF model.\\

\hline
\begin{tabular}{l}LEGAL-BERT-BiLSTM-CRF~\cite{article_1}\end{tabular} 
& Uses a LEGAL-BERT model for generating input sentence embeddings, which are then used with a BiLSTM-CRF model.\\

\hline
\begin{tabular}{l}sciBERT-HSLN~\cite{DBLP:journals/corr/abs-2201-13125}\end{tabular} 
& Uses BERT embeddings with a BiLSTM-attention-pooling-CRF model.\\

\hline
\begin{tabular}{l}MTL (BERT-SC)~\cite{DBLP:journals/corr/abs-2112-01836}\end{tabular} 
& Uses a BiLSTM-CRF model in a multitask setting. The auxiliary task is label shift prediction, for which a pretrained BERT is finetuned initially for binary classification. The shift embeddings generated from this model, are then fed into a BiLSTM-CRF which works with another model trained for rhetorical role classification.\\
\hline 
\begin{tabular}{l}Gemini-pro \end{tabular} 
& We leverage Gemini-pro\footnote{https://ai.google.dev/models/gemini} model for labeling the rhetorical rule of the documents and explore both zero-shot and few-shot approaches.\\
\hline

\multicolumn{2}{|c|}{{\bf MARRO family of models}} \\ \hline

MARRO\textsubscript{base} (This work) 
& Uses a pretrained sent2vec model to generate initial sentence embeddings, over which multi-headed attention is applied, before being fed into a BiLSTM-CRF model.\\

\hline
TF-MARRO (This work) 
& Uses a legal-BERT-SMALL model to generate sentence embeddings, over which multi-headed attention is applied, before being fed into a BiLSTM-CRF model. The legal-BERT-SMALL model is not pretrained but is constantly fine-tuned during TF-MARRO training, which leads to more relevant sentence embeddings.\\

\hline
MTL-MARRO (This work) 
& Trains a MARRO model in a multitask setting, using label shift as an auxiliary task. Unlike MTL (BERT-SC), we use an off-the-shelf BERT model for shift embeddings, which is fine-tuned at the same time as the rhetorical role model.\\

\hline
MTL-TF-MARRO (This work) 
& Trains a TF-MARRO in a multitask setting as defined in the above row.  \\
\hline
\end{tabular}}
\caption{Brief descriptions of baseline models and how they compare with the MARRO family of models proposed in this work.}
\label{table:models_desc}
\end{table}

\textbf{MTL-TF-MARRO}: This variant of TF-MARRO uses LEGAL-BERT-SMALL for generating the input embeddings, for the rhetorical role prediction module. We experimented with different transformer models for input embeddings, BERT, LEGAL-BERT, and LEGAL-BERT-SMALL to name a few. We noticed using a smaller
 version of these models (LEGAL-BERT-SMALL) produces faster training with no decrease in performance. Using bigger models like BERT takes significantly more resources because a:) They are slower for inferencing than their distilled counterparts, b:) Since the input embeddings model is not used off the shelf, but fine-tuned as the training progresses, backpropagation and gradient update is faster for smaller models.\\ 

Both these models (MTL-MARRO and MTL-TF-MARRO) remove the need to have a separate BERT/LEGAL-BERT model for the auxiliary task and rhetorical role classification task. Further, using  the single trainable model and fully connected layers in the label shift prediction side ensures embeddings account for gradient update from the rhetorical role classifier model as well (as compared to static embeddings used in~\cite{DBLP:journals/corr/abs-2112-01836})

\section{Experimental Evaluation}
\noindent This section discusses the experiments we performed and the analysis of obtained results from various models. To have a standardized comparison with results from the previously proposed models, we evaluate the rhetorical role labeling performance using a five-fold cross-validation. We also study the label-wise performance of the models for each dataset. Finally, this section shows some actual sentences from both the datasets, their gold standard labels done by the annotators, and how our proposed model(s) MARRO performs as compared to previously available models.

\subsection{Experimental Setup}
\noindent To evaluate the effectiveness of proposed models, we perform rhetorical role classification on the two datasets explained in Section~\ref{sec:dataset}.  
Each document is a separate input to the model, hence the initial embeddings of each sentence are only a function of words in the current sentence and the sentences in the current document. We train the models using Google Colab on NVIDIA Tesla K80 GPUs with 5-fold cross-validation, trained for 80 epochs. 
Training for all five folds takes 2.5 to 3 hours, per dataset. We also note that the $D_{UK}$ dataset regularly has documents with more than 500 sentences, so the models can hold up well with lengthy documents with the training time still being manageable with the above-stated free-of-cost resources. Once the training is done, the actual inferencing of documents is much faster. On average, generating rhetorical roles for a document from our dataset takes between 1.1 to 1.2 seconds, when using a T4 GPU on google colab.
The evaluation metric is \textit{macro-averaged F1 score, averaged across all 5 folds}. The embedding dimension is 512 for the TF-MARRO model and 200 for the MARRO\textsubscript{base} model. For the Gemini-pro model, hyperparameters details are temperature=0, top-k=1, and top-p=0.1. \\

\noindent The learning rate of both the models is set to 0.1 for $D_{IN}$ dataset and 0.001 for $D_{UK}$dataset. The embedding dimension is 512 for the LEGAL-BERT-SMALL model (used in TF-MARRO and MTL-TF-MARRO) and 200 for the sent2vec model (used for embeddings in MARRO\textsubscript{base} and MTL-MARRO). For the multi-headed attention layer, we found that models performed best when each head in the attention encoder was working on feature dimensions in the range of 32 to 64. Hence, for the pretrained embedding-based models (MARRO\textsubscript{base} and MTL-MARRO) we use 5 attention heads (200/5 = 40). For transformer embedding-based models (TF-MARRO and MTL-TF-MARRO) we use 8 attention heads (512/8 = 64). Multi-headed attention is usually implemented in n \begin{math}\in\end{math} \{6,12\} encoder blocks, where the learning rate is 0.0001 and epoch size is 50. We have chosen 2 encoder blocks since that was the best trade-off between performance and model size/ training speed. \SA{The BiLSTM-CRF takes the input from the attention layer, where the output from the label shift module BiLSTM is concatenated with the BiLSTM output of the rhetorical role prediction module, before being fed into the CRF, as discussed above.}. For LEGAL-BERT-SMALL, we have kept the last 2 layers as trainable and the sent2vec model is pretrained. The sent2vec model does not need to be fine-tuned along with the model. The reason is the original 53,000 documents that were used to create this model were from the Indian Supreme Court and hence the embeddings from this model closely resemble the dataset from the Indian Supreme Court we have. This is not the case with the UK dataset, where we found out the LEGAL-BERT-SMALL fine-tuned embeddings gave better contextual embeddings than a pretrianed sent2vec. We compare our proposed models against the baseline models stated in Section~\ref{sub:baselines}.
\begin{table*}[!htb]
\centering
\resizebox{0.8\textwidth}{!}{
\begin{tabular}{||p{2.6cm}|p{3.2cm}||p{0.9cm}|p{0.9cm}|p{0.9cm}||p{0.9cm}|p{0.9cm}|p{0.9cm}||}
\hline

\multicolumn{1}{||c|}{\bfseries Model} & \multicolumn{1}{c||}{\bfseries Variant} & \multicolumn{3}{c||}{\bfseries D\textsubscript{IN}} & \multicolumn{3}{c||}{\bfseries D\textsubscript{UK}} \\ \hline
&& {P} & {R} & { F1} & {P} & {R} & {F1}\\ \hline

\multicolumn{8}{|c|}{Baselines} \\ \hline
\raggedright
SciBERT-HSLN \cite{DBLP:journals/corr/abs-2201-13125} &  Base version &  0.689 &  0.581 &  0.609 &  0.567 &  0.444 &  0.444 \\
\break
\raggedright
Hierarchical BiLSTM-CRF \cite{bhattacharya2019identification} &  Pretrained embeddings &  0.729 &  0.655 &  0.683 &  0.533 & 0.476 &  0.491 \\
\break
BERT-BiLSTM-CRF \cite{article_1} &  2 Layers of BERT-base fine tuned & 0.698 &  0.645 & 0.667 & 0.653 &  0.567 & 0.580 \\
\break
LEGAL-BERT-BiLSTM-CRF \cite{article_1} & 2 Layers of LEGAL-BERT-base fine tuned &  0.692 &  0.639 &  0.661 &  0.653 &  0.587 &  0.600 \\
\break
\raggedright
MTL (BERT-SC) \cite{DBLP:journals/corr/abs-2112-01836} &  Base version &  0.735 &  0.673 &  0.692 & \textbf{0.657} &  0.569 &  0.594 \\
\break
\raggedright
Gemini-pro & Zero-shot & 0.481 & 0.493 & 0.377 & 0.312 & 0.492 & 0.248  \\
\break
\raggedright
Gemini-pro & One-shot & 0.463 & 0.524 & 0.435 & 0.332 & 0.544 & 0.285\\
\hline

\multicolumn{8}{|c|}{ Proposed models (all variants with 2 encoder blocks)} \\ \hline
\break
MARRO\textsubscript{base} & 5 att-heads per block &  0.768 & 0.668 & 0.706 & 0.526 & 0.495 & 0.502 \\
\break
TF-MARRO & 8 att-heads per block &  0.700 & 0.660 &  0.675 &  0.640 & 0.594 & 0.608 \\
\break
MTL-MARRO & 5 att-heads per block &  \textbf{0.787} &  \textbf{0.685} &  \textbf{0.724} &  0.609 &  0.535 &  0.552 \\
\break
 MTL-TF-MARRO  & 8 att-heads per block &  0.712 &  0.615 &  0.657 &  0.652 &  \textbf{0.600} &  \textbf{0.617} \\
\hline
\end{tabular}}
\caption{Results of rhetorical role labeling on both datasets. The evaluation metric is a macro-averaged F1 score across a 5-fold cross-validation dataset. Precision (P) and Recall (R) values are also shown. The best values are shown in bold. We also conducted a two-tailed t-test~\cite{article_t_test} on the F1 scores with a 0.05 significance level. We propose the null hypothesis ($H_0$) that says the results are not significant for our model. The alternate hypothesis ($H_a$) we take is the results (F1) from our model are statistically significant. For the $D_{IN}$, we got a p-value of 0.0013, which is lesser than 0.05, hence we rejected the ($H_0$), concluding our results from the proposed model (MTL-MARRO) are significantly better than the previous best (MTL (BERT-SC)). For the $D_{UK}$ dataset, we got a p-value of 0.0891, which is greater than 0.05, hence we can't reject the null hypothesis. \RE{This indicates that there is insufficient evidence to support the alternative hypothesis (Ha). Consequently, the results of our model (MTL-TF-MARRO) do not show a statistically significant improvement over the previous best model (LEGAL-BERT-BiLSTM-CRF).} }
\label{table:results_all_models}
\end{table*}


\subsection{Results}
\noindent \textbf{Overall Performance Analysis:} The results from the experiments are shown in Table \ref{table:results_all_models}. The proposed multi-task models with multi-headed attention outperform previous state-of-the-art models. 
On the $D_{IN}$ dataset, the best performing model was MTL-MARRO with an F1 of 0.724, followed by MARRO\textsubscript{base} (model trained without an auxiliary task). We see that for this dataset, models using pretrained sent2vec embeddings perform considerably better than BERT or LEGAL-BERT embedding models. That is because the LEGAL-BERT model was trained mainly on EU, UK, and US legal documents, and the terminology is considerably different for these documents as compared to their Indian counterparts. The superior performance of the LEGAL-BERT-based embedding model is corroborated in the findings of \cite{article_1}.

For the $D_{UK}$ dataset, the best-performing model was MTL-TF-MARRO, which produced an F1 score of 0.617, marginally better than LEGAL-BERT-BiLSTM-CRF (0.600). \RE{Since there is an overlap in the vocabulary and semantics of this dataset and the pretraining corpora of the LEGAL-BERT-SMALL model, the BERT/LEGAL-BERT-SMALL embedding models perform much better as compared to using fixed sent2vec embeddings here. Although it is difficult to establish whether there was an explicit overlap of these datasets with our data, based on performance of our model, we hypothesize that familiarity with the language and legal terminologies could be a reason why models perform better with LEGAL-BERT-SMALL embeddings for the UK Supreme Court datasets.} It is to be noted that LEGAL-BERT-SMALL embeddings do not perform well with the D\textsubscript{IN} dataset, probably because this model was not pretrained on documents from the Indian Judiciary. 
Hence, we get better results by using MARRO\textsubscript{base} for the D\textsubscript{IN} dataset and the TF-MARRO model for the D\textsubscript{UK} dataset. Our models outperform the previous multi-task learning models suggested by Malik et al. \cite{DBLP:journals/corr/abs-2112-01836}. \\

\begin{table*}[t]
\centering
\resizebox{1\textwidth}{!}{
\begin{tabular}{||p{0.9cm}||p{1.4cm}|p{1.4cm}|p{1.4cm}|p{1.4cm}|p{1.4cm}|p{1.4cm}||p{1.4cm}|p{1.4cm}|p{1.4cm}|p{1.4cm}|p{1.4cm}|p{1.4cm}||}
\hline
\multicolumn{1}{||c||}{\bfseries Role} & \multicolumn{6}{|c||}{\bfseries D\textsubscript{IN}} & \multicolumn{6}{|c||}{\bfseries D\textsubscript{UK}} \\ \hline
& 
\raggedright
{ Hier. BiLSTM-CRF \cite{bhattacharya2019identification}} 
\raggedright
& { LEGAL-BERT-BiLSTM-CRF \cite{article_1} }
\raggedright
&  {MTL (BERT-SC) \cite{DBLP:journals/corr/abs-2112-01836} } 
\raggedright
& {Gemini-pro (Zero-shot)}
\raggedright
&{Gemini-pro (Few-shot)}
& {  MTL-MARRO (proposed) }
\raggedright
& {  Hier. BiLSTM-CRF \cite{bhattacharya2019identification}} 
\raggedright
& {  LEGAL-BERT-BiLSTM-CRF \cite{article_1}}
\raggedright
&  {  MTL (BERT-SC) \cite{DBLP:journals/corr/abs-2112-01836}} 
\raggedright
&{Gemini-pro (Zero-shot)}
\raggedright
&{Gemini-pro (Few-shot)}
\raggedright
& { MTL-TF-MARRO (proposed)} \\
\hline
FAC & 0.828 & 0.789 & 0.790 &0.577  &0.632 &\textbf{0.839} & 0.719 & 0.778 & 0.757 &0.472&0.441& \textbf{0.779}\\
\hline
ARG & 0.492 & \textbf{0.632} & 0.521 &0.528 &0.524 & 0.573  & 0.147 &  0.393 & 0.356 &0.233&0.182& \textbf{0.430}  \\
\hline
RATIO & 0.860 & 0.833 & \textbf{0.890} &0.316 &0.335& 0.876& 0.815 & 0.838 & 0.826 &0.311&0.432& \textbf{0.843}\\
\hline
STA & 0.657 & 0.700 & 0.668 &0.548 &0.577 & \textbf{0.712} & 0.552 & 0.617 & 0.642 & 0.294&0.310&\textbf{0.663}\\
\hline
PRE &  0.526 & 0.584 & \textbf{0.661} &0.282 &0.315 & 0.568 & 0.377 &0.479  & 0.469 & 0.320&0.370&\textbf{0.487} \\
\hline
RPC & 0.831 & 0.761  & 0.814 &0.241 &0.356& \textbf{0.857}  & 0.551 & 0.603  & 0.628 &0.051 &0.112& \textbf{0.647}\\
\hline
RLC & 0.590 & 0.316 &  0.498 &0.147 &0.312 &\textbf{0.641} & 0.270 &  0.412  & \textbf{0.475} &0.060 &0.153& 0.469 \\
\hline
\end{tabular}}
\caption{
F1 scores of different models on the classification of each rhetorical role on both datasets. Label-wise best F1 scores are highlighted in bold.}
\label{table:label_wise_f1}


\end{table*}

\noindent

\textbf{Label-wise Performance Analysis : } As shown in Table \ref{table:label_wise_f1}, for the $D_{IN}$ dataset, our model beats the previous SOTA on 4 out of 7 rhetorical roles. 
\RE{Both the previous SOTA and our model perform poorly on \emph{Argument} labels because arguments usually occur between sentences of other label types, although the labels within a single sentence or between consecutive sentences usually do not change. \citet{DBLP:journals/corr/abs-2112-01836} suggest that 88\% of the times, the labels stay the same. However, It is possible that the same label occurs at multiple places in the document; that is, while the labels occur in contiguous blocks, that is not always the case.} For the $D_{UK}$ dataset, our proposed model improves on 6 out of 7 rhetorical roles, with only a marginal drop in the RLC label. Similar to the Indian dataset, the Argument label is predicted poorly, while other labels like FAC and Ratio are predicted with comparatively higher accuracy. The models perform poorly on $D_{UK}$ as compared to $D_{IN}$ data on RLC and RPC labels, the reason being that in the UK, each judge or ``lord'' writes a separate paragraph for their case judgment. This makes it difficult for the model to learn the semantic dependency of this label in UK legal documents. For both the datasets, our model performs better on most of the labels as compared to MTL (BERT-SC) \cite{DBLP:journals/corr/abs-2112-01836} (except for the label PRE in $D_{IN}$ where MTL (BERT-SC) outperforms our model). This goes on to show the effectiveness of proposed multi-headed attention and trainable label shift embeddings. \\ 

In Table~\ref{table:results_all_models}, we showcase the overall result of gemini-pro model. In the few-shot setting, the average F1 score is 0.432 and 0.281 for the $D_{IN}$ and $D_{UK}$ respectively, which is slightly better than the zero-shot approach. Furthermore, in the class-wise result, we witness that RPC and RLC yield low scores in both approaches, as presented in Table~\ref{table:label_wise_f1}.\\

\noindent To test the significance of proposed model(s), we conduct a two-tailed t-test ~\cite{article_t_test} on the F1 scores of the models with 0.05 significance level. We find the results of our proposed model (MTL-MARRO) are significantly better than the previous best (MTL (BERT-SC)) with a p-value of 0.0013, for the $D_{IN}$ dataset. 
For the $D_{UK}$ dataset, the results of the proposed model are not significant against the previous best (LEGAL-BERT-BiLSTM-CRF) with a p-value of 0.0891.\\

\noindent Label-wise agreement analysis for the gold standard labels (human annotators) and the outputs of the best models shows that label ARG is frequently misclassified as Ratio in both the datasets. Further, the label RPC has rarely been misclassified in the  $D_{IN}$  dataset since it always occurs at the end of the document, while in the  $D_{UK}$  dataset, it is often misclassified as Ratio for reasons mentioned in the preceding paragraph. We also find that sentences labeled PRE are often classified as RATIO by the models in both datasets. This is because the RATIO (rationale) involves prior cases and there is interleaving between sentences of these types. This disagreement also exists among legal experts, showing that these labels are subjective. We have provided a sentence-wise comparison of all rhetorical roles for select sentences, and how the previous and our proposed models vary in results for them. 

\noindent Table~\ref{table:docs_india} and Table~\ref{table:docs_uk} show examples of sentences from both the datasets and performance of baseline models and the best performing proposed models (MTL-MARRO and MTL-TF-MARRO). For each label, we exhibit 2 sentence pairs, one where the proposed models give the correct prediction (often improving over the incorrect predictions made by the baseline models) and another where the proposed models fail to give the correct results.

\noindent For the $D_{IN}$ dataset, both the proposed model (MTL-MARRO) and the previous state-of-the-art often confuse the sentences with label Ratio with FAC (and vice-versa), as shown in the examples below. The proposed model generally performs better, proving more robust in predicting label changes, thanks to the label shift auxiliary task. This is shown when the shift from labels FAC to ARG is predicted more accurately by MTL-MARRO, while the previous state-of-the-art model might predict the same label as the previous sentence, before detecting the label change and then predicting the correct label in the following sentences. 

\noindent For the $D_{UK}$ dataset, both the proposed model (MTL-TF-MARRO) and the previous state-of-the-art confuse FAC with Ratio, while ARG labels occurring immediately after a FAC label are harder to predict. Label PRE is mostly incorrectly predicted to be Ratio, in both the models.

\begin{table}[h]
    \centering
    \resizebox{0.8\textwidth}{!}{
    \begin{tabular}{||p{7cm} p{1cm} p{1.2cm} p{1.2cm} p{1.2cm}||}
    \hline
         \centering \textbf{Sentence Examples} & \textbf{Gold Labels} & \textbf{Hier BiLSTM-CRF} & \textbf{MTL-MARRO} & \textbf{Gemini-pro (Few-shot)} \\
         \hline \hline
         It is urged that the Government of India would have the same defences ... constitutional limitations.& ARG & ARG & ARG & ARG \\
         
         We are of opinion that the submission of the learned Attorney General is correct.& \textbf{Ratio} & ARG & \textbf{Ratio} & RPC\\
         \hline \hline
         The High Court however gave certificates to the Union of India ... two appeals before us.&\textbf{Ratio} & FAC & FAC &  RPC \\ 
         One against the decree passed in the suit ...  in the two appeals are exactly the same. & \textbf{Ratio} & FAC & FAC & RPC\\
\hline \hline
3) The Commissioner shall ... the case, may grant such relief, if any, as it thinks appropriate. & \textbf{FAC} & STA & \textbf{FAC} & RPC \\  
In December 1950, the company applied ...  for concessions regarding income tax and super tax. & FAC & FAC & FAC & STA\\
\hline \hline
 Three main contentions were raised on behalf of the company in the High Court. & \textbf{FAC}& \textbf{FAC} & Ratio & ARG\\  
In the first place it was urged that the order dated January 18, 1947 was a special law. &\textbf{FAC} &\textbf{FAC} & Ratio & ARG\\
\hline \hline
This being the case, it is argued, ... Ground No. 1 was, in fact, argued or not before the Tribunal. & \textbf{ARG} & FAC & FAC & \textbf{ARG}\\  
According to Mr.Sampath, over 5 years' delay in  ... seeking relief on the basis of Ground No. 1. & \textbf{ARG} & FAC &\textbf{ARG} &\textbf{ARG}\\
\hline \hline
It has spread everywhere. & \textbf{PRE} &ARG &\textbf{PRE} & RPC  \\  
No facet of public activity has been left unaffected by the stink of corruption.&\textbf{PRE} & ARG &\textbf{PRE} & \textbf{PRE} \\
\hline \hline
Though,these grounds may be attractive ... the same cannot be applied to the case on hand .. & Ratio & Ratio & Ratio & Ratio\\  
About the request based on delay that the appellant ... refer decision of this Court in State of M.P.vs .. &\textbf{PRE} &Ratio & Ratio &\textbf{PRE}\\
\hline \hline
 Sub section(4) of section 7 of the Repealing Act ... declared by law to be evacuee property". & \textbf{STA} &Ratio & \textbf{STA} &\textbf{STA} \\  
Reference may also be made ... of the setting up of the Dominions of India and Pakistan.&\textbf{STA}&Ratio&\textbf{STA} &\textbf{STA} \\
\hline \hline
In the result, the writ petitions are disposed of subject to the observations made above. 
&\textbf{RPC} & Ratio & \textbf{RPC} & \textbf{RPC}\\  
R.S.S. Petitions disposed of. &\textbf{RPC} &\textbf{RPC}&\textbf{RPC} & \textbf{RPC}\\
\hline \hline
 We, therefore, confirm till further orders the interim order made by us on April 20, 1981 & \textbf{RLC} & \textbf{RLC} &\textbf{RLC} & RPC\\  
The terms of the said order, that is on the undertaking given on behalf ... equalization account &\textbf{RLC} & Ratio &\textbf{RLC} & RPC \\
\hline \hline
Petitions were then filed by Banarasi Prasad and the transferees  ... improper and ulterior motives. & FAC & FAC & FAC & ARG \\  
The High Court rejected these petitions ...  was to file suits for relief in the civil court. & \textbf{RLC} & FAC & FAC & \textbf{RLC} \\
\hline \hline
\end{tabular}}
    \caption{Label-wise Prediction Analysis from Indian Court Documents. The second column shows the gold standard labels, and the last two columns show the predictions by the Hierarchical BiLSTM-CRF and the MTL-MARRO models respectively.}
    \label{table:docs_india}
\end{table}

\begin{table}[t]
    \centering
    \resizebox{0.8\textwidth}{!}{
    \begin{tabular}{||p{7cm} p{1cm} p{1.2cm} p{1.2cm} p{1.2cm}||}
    \hline
         \centering \textbf{Sentence Examples} & \textbf{Gold Labels} & \textbf{Hier BiLSTM-CRF} & \textbf{MTL-MARRO} & \textbf{Gemini-pro (Few-shot)}\\
         \hline \hline
         With respect to certain eminent judges ...  appended to his decision was entirely misconceived.  & FAC & FAC & FAC & RPC\\  
It was that, because the Balen report was outside the designation ... Act in relation to Mr Sugars request. & \textbf{FAC} & ARG & \textbf{FAC} & ARG \\
\hline \hline
 Concerned that he was being invited ... constructions which the Tribunal had rejected.  
 & FAC & FAC & FAC & Ratio \\  
In the event he adopted the BBCs polarised construction. & \textbf{FAC} &\textbf{FAC} & ARG & ARG \\
\hline \hline
 The statutory position in Northern Ireland & \textbf{STA} & RLC & \textbf{STA} & \textbf{STA} \\ 
Article 6 Table A of the Rehabilitation of Offenders ... the expiry of five years. &\textbf{STA} & RLC & \textbf{STA} & \textbf{STA}\\
\hline \hline
In this case however we are concerned with meaning and effect of the statute. &\textbf{STA} & RPC & Ratio  & \textbf{STA} \\  
The relevant provisions are to be ... Asylum Act 2002, which deals with immigration and asylum appeals. & \textbf{STA} & RPC & Ratio & \textbf{STA} \\
\hline \hline
It was to be interpreted, both ... law and in a sensible and practical rather than a purely literal way. &\textbf{RLC} & FAC & \textbf{RLC} & Ratio \\  
On the facts of Ms Agyarkos case, the Secretary of States conclusion ... together overseas. &\textbf{RLC} & FAC & \textbf{RLC} & RPC\\
\hline \hline
The mere facts that ... could not constitute insurmountable obstacles to his doing so. &\textbf{RLC} & \textbf{RLC} & FAC & FAC \\  
(para 25) On the facts of Ms Ikugas case, Sales LJ agreed with ... (para 50).& \textbf{RLC} & \textbf{RLC} & FAC & RPC \\
\hline \hline
In this judgment, PAC submits, the CJEU refined its jurisprudence ... from overseas. &\textbf{ARG}&\textbf{ARG}&\textbf{ARG} & RPC \\  
HMRC in reply point out that FII ECJ II was concerned ... its treatment of Haribo.&\textbf{ARG}& Ratio &\textbf{ARG} & \textbf{ARG} \\
\hline \hline
 In response, PAC relies upon a later CJEU decision  ... EU:C:2012:707; [2013] Ch 431 (FII ECJ II). &\textbf{ARG} &\textbf{ARG}&\textbf{ARG} & PRE\\  
Judge Lenaerts was once again the juge rapporteur.&\textbf{ARG}&\textbf{ARG}& Ratio & RPC \\
\hline \hline
In assessing what would have been a fair share  ... out in section 41(4).&\textbf{Ratio}& FAC & \textbf{Ratio} &\textbf{Ratio} \\  
In so doing he had regard to the nature of Professor Shanks duties ... but not exceptional. &\textbf{Ratio}& FAC & \textbf{Ratio} & FAC \\
\hline \hline
True it is that section 41(2) does not ... interpreted in that way.&\textbf{Ratio}& \textbf{Ratio}& ARG & \textbf{Ratio} \\  
This is still not a complete answer, however, for the deeming ... patents in issue may be judged.&\textbf{Ratio}& \textbf{Ratio}& ARG & FAC\\
\hline \hline
It will not always be clear what those purposes are. &\textbf{PRE} &Ratio&\textbf{PRE} & Ratio \\  
If the application of the provision would ... it should not be applied.&\textbf{PRE}&Ratio&\textbf{PRE} & Ratio \\
\hline \hline
The facts were striking in that, had the patents not existed, ... its business was transformed.&\textbf{PRE}&\textbf{PRE}&Ratio & FAC \\  
The commercial embodiment of the invention of the patents was ... total value in excess of 1.3 billion.&\textbf{PRE}&\textbf{PRE}&Ratio & FAC\\
\hline \hline
\end{tabular}}
    \caption{Label-wise Prediction Analysis from UK Court Documents. The second column shows the gold standard labels, and the last two columns show the predictions by the Hierarchical BiLSTM-CRF and the MTL-MARRO models respectively.}
    \label{table:docs_uk}
\end{table}

\subsection{Bottlenecks and Optimisation}
\noindent In this section, we discuss the potential bottlenecks in training and inferencing MARRO and how can the performance be optimized. Firstly, the architecture relies heavily on sentence embeddings, either from the sent2vec model or the LEGAL-BERT-SMALL model. For the MARRO variant using sent2vec, embeddings can be generated beforehand, once per dataset, and hence don't affect training time. For the MARRO variant that uses LEGAL-BERT-SMALL, each sentence is passed through the LEGAL-BERT-SMALL model at the runtime to produce the embeddings, which has a significant impact on the training time. With our experience, this can be mitigated using a larger batch size or parallel sentence processing (using threads or multi-processing). This is possible because the embeddings for one sentence are independent of others; the actual contextual sentence embeddings are obtained later in the model when we explicitly employ multi-headed attention. Apart from this, we also fine-tune the LEGAL-BERT-SMALL model used in the label shift prediction task during the training process, which adds to training time as well. This can be changed and we can fine-tune the model beforehand much like ~\cite{DBLP:journals/corr/abs-2112-01836}. While this might take some time off training, we found the results to be better when the auxiliary task LEGAL-BERT-SMALL was fine-tuned during the training and not beforehand. Since the overall training time is manageable and under 3 hours, we decided this was a fair tradeoff to improve performance. 


\section{Conclusion}
\noindent AI has the potential to herald unprecedented contributions to legal analysis. With the onset of digitization, automated document parsing, and understanding, stakeholders in the judiciary, researchers, and society stand a lot to gain in terms of improving the legal workflow and consequently faster justice deliverance. It is hence important to define clearly the theoretical and societal implications of our work. On the research and theoretical front, we have discussed and explored how the use of AI and NLP can aid in better rhetorical role identification. For our problem, we have established a hybrid solution employing both LSTM and transformer models to give the best results. Our findings are in line with previous works \cite{bhattacharya2019identification, article_1, DBLP:journals/corr/abs-2112-01836} that state BiLSTM-CRF is a good base model for rhetorical role labeling and specialized embeddings render better results for the task. We however find trainable embeddings are better when training a model in a multi-task setting, unlike \cite{malik-etal-2021-ildc}. Also, we explicitly experiment with the placement of multi-headed attention layers to improve upon the previous models. This brings forth a research implication that could potentially steer the direction of legal rhetorical role labeling from previously used BiLSTM-CRF models to attention-based BiLSTM-CRF models trained in a multi-task setting. \\
On the societal front, our work implies that legal AI can be an indispensable tool for aiding lawyers in faster rhetorical role labeling. Currently, most of this labeling is done manually, and having a reliable legal rhetorical role labeling software would mean lawyers would have more time to work and study the case, leading to faster preparation and reduced case trial times. In the long run, as these AI systems become more reliable requiring lesser human intervention, this can greatly expedite court rulings and help alleviate pending cases in the judiciary.  
\noindent We propose a novel family of models called MARRO for rhetorical role labeling, with four models (MARRO\textsubscript{base}, TF-MARRO, MTL-MARRO, and MTL-TF-MARRO) that outperform the previous state-of-the-art on 2 datasets, from India and the UK respectively. 
We also develop a bigger rhetorically labeled dataset of legal documents from India ($D_{IN})$. The multi-headed attention layer enriches the input embeddings obtained from a sent2vec or LEGAL-BERT-SMALL model, with neighboring sentences' context. These embeddings are used with a BiLSTM-CRF model, that has been previously shown to be effective for the rhetorical role identification task. We further train the said models in a multi-task learning setup with label shift prediction as an auxiliary task. The dataset and our model implementations will be made public upon acceptance of the paper. 

\SA{However, there are still significant bottlenecks in large-scale adaptation of such tools in the judicial landscape. \citet{girard2023travail} delineates automatic pseudonymization, which is an important research direction we did not explore in our current work. Further, explainability is especially difficult to achieve for legal AI, because unlike explainable AI for other tasks, here explicit reference to authoritative legal sources are to be made for a system to be considered reliable~\cite{branting2019semi}}. Previous works and our findings show predicting rhetorical roles correctly is harder than other multi-label classification problems since even experts differ in defining the gold standard labels. Also, the performance depends heavily on the nature of the embeddings and how they were generated, hence there is no single model that can work on legal documents from all countries. In our case LEGAL-BERT embeddings based MARRO worked better for the UK dataset, while pretrained sent2vec embeddings MARRO worked better for the Indian dataset. While vast improvements have been in using AI for rhetorical role labeling, the bottleneck remains the dearth of annotated datasets. Our work is a simultaneous attempt at expanding the available datasets and showing the effectiveness of the model and embedding selection to obtain better results than previous works. 

Future work would involve minimizing dependence on the source of embeddings such that one model can work with legal documents from any country, provided ample training data is present. Thereafter, we look forward to building a user-friendly implementation of the same with expert guidance. It would be a good future project to create a toolkit in terms of a pip package or a library so legal practitioners can benefit from our model. Identification of rhetorical roles is a powerful tool that can assist other important tasks that currently require manual effort. In the future, we also hope to extend this work to improve downstream tasks like legal document summarization and judgment prediction.

\end{document}